\documentclass[letterpaper]{article} 
\usepackage{aaai23}  
\usepackage{times}  
\usepackage{helvet}  
\usepackage{courier}  
\usepackage[hyphens]{url}  
\usepackage{graphicx} 
\usepackage{natbib}  
\usepackage{caption} 
\usepackage{algorithm}

\usepackage{newfloat}
\usepackage{listings}
\DeclareCaptionStyle{ruled}{labelfont=normalfont,labelsep=colon,strut=off} 
\floatstyle{ruled}
\newfloat{listing}{tb}{lst}{}
\floatname{listing}{Listing}
\usepackage{booktabs} 
\usepackage{amssymb}
\usepackage{amsmath,bm}
\usepackage{multirow}

\nocopyright 

\title{Bi-directional Feature Reconstruction Network for Fine-Grained \\ Few-Shot Image Classification}
\author {
    Jijie Wu\textsuperscript{\rm 1}, 
    Dongliang Chang\textsuperscript{\rm 2},
    Aneeshan Sain\textsuperscript{\rm 3},\\
    Xiaoxu Li\textsuperscript{\rm 1}\thanks{indicates corresponding author.},
    Zhanyu Ma\textsuperscript{\rm 2},
    Jie Cao\textsuperscript{\rm 1},
    Jun Guo\textsuperscript{\rm 2}, 
    and Yi-Zhe Song\textsuperscript{\rm 3}
}
\affiliations {
    \textsuperscript{\rm 1}School of Computer and Communications, Lanzhou University of Technology, Lanzhou, China\\
    \textsuperscript{\rm 2}School of Artificial Intelligence, Beijing University of Posts and Telecommunications, Beijing, China\\
    \textsuperscript{\rm 3}SketchX, CVSSP, University of Surrey, United Kingdom \\
    
    \{jijie, lixiaoxu, caoj\}@lut.edu.cn, 
    \{changdongliang, mazhanyu, guojun\}@bupt.edu.cn, 
    \{a.sain, y.song\}@surrey.ac.uk
}

\begin{document}

\maketitle

\begin{abstract}
The main challenge for fine-grained few-shot image classification is to learn feature representations with higher inter-class and lower intra-class variations, with a mere few labelled samples. Conventional few-shot learning methods however cannot be naively adopted for this fine-grained setting -- a quick pilot study reveals that they in fact push for the opposite (i.e., lower inter-class variations and higher intra-class variations). To alleviate this problem, prior works predominately use a support set to reconstruct the query image and then utilize metric learning to determine its category. Upon careful inspection, we further reveal that such unidirectional reconstruction methods only help to increase inter-class variations and are not effective in tackling intra-class variations. In this paper, we for the first time introduce a bi-reconstruction mechanism that can simultaneously accommodate for inter-class and intra-class variations. In addition to using the support set to reconstruct the query set for increasing inter-class variations, we further use the query set to reconstruct the support set for reducing intra-class variations. This design effectively helps the model to explore more subtle and discriminative features which is key for the fine-grained problem in hand. Furthermore, we also construct a self-reconstruction module to work alongside the bi-directional module to make the features even more discriminative. Experimental results on three widely used fine-grained image classification datasets consistently show considerable improvements compared with other methods. 
Codes are available at: https://github.com/PRIS-CV/Bi-FRN.
\end{abstract}

\section{Introduction}

\begin{figure}[!htp]
  \centering
  \includegraphics[width=0.85\linewidth]{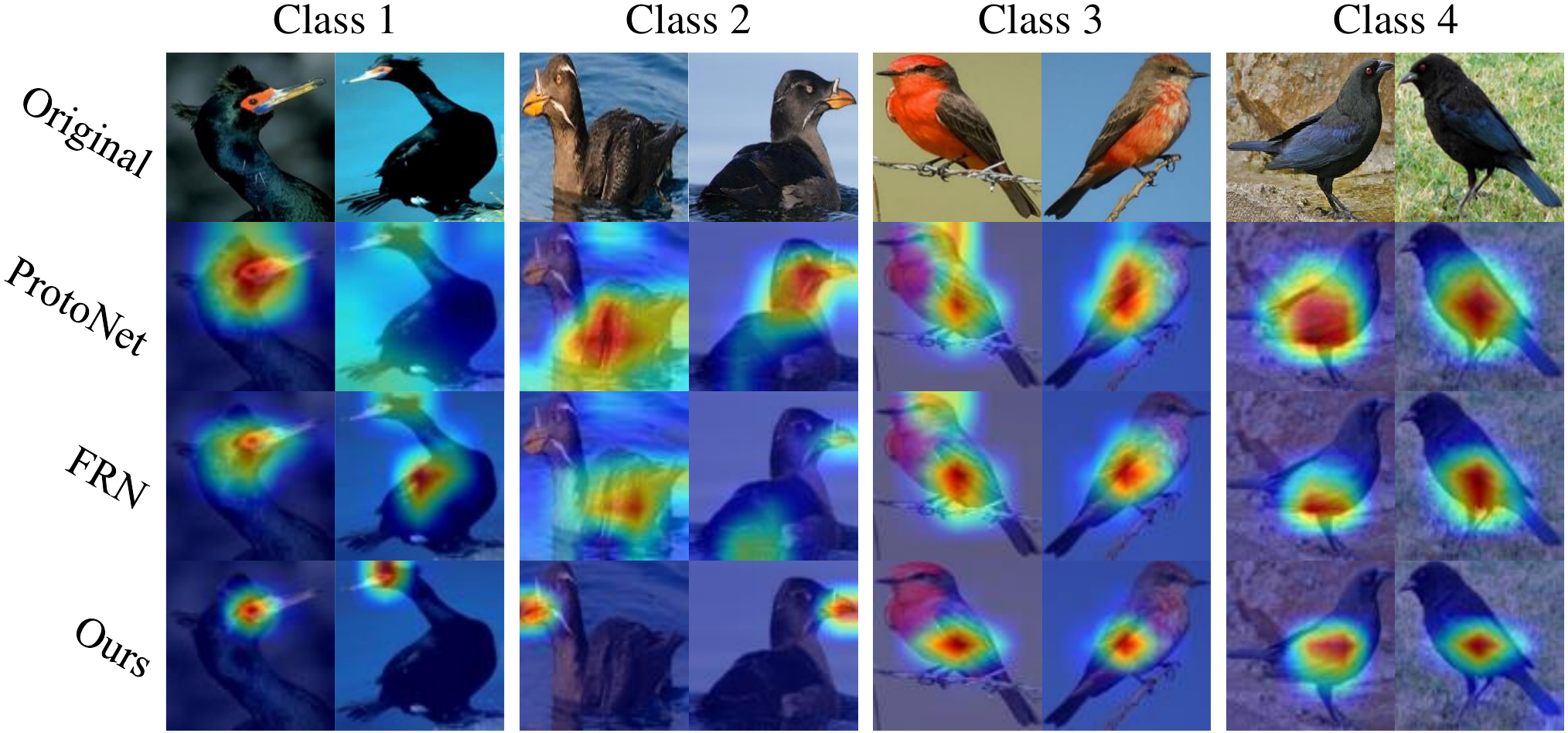}
  \caption{Visualization of the localized regions returned from Eigen-CAM~\cite{muhammad2020eigen} based on a ProtoNet~\cite{NIPS2017_cb8da676} model (trained on CUB-200-211 dataset) optimized by the traditional method, the FRN~\cite{Wertheimer_2021_CVPR}, and the proposed method. The higher energy region denotes the more discriminative part in the image.}
  \label{fig:1}
\end{figure}

Few-shot fine-grained image classification~\cite{Huang2021LowRankPA,9293172} has emerged very recently as an important means to tackle the data scarcity problem widely facing fine-grained analysis~\cite{3401454f29a841a29142c97833857e78}. As per the conventional few-shot setting, it asks for effective model transfer given just a few support samples. The differences however lies in the unique characteristics brought by the fine-grained nature of the problem -- the model needs to focus on learning subtle and discriminative features to discriminate not only on the category level (as per conventional few-shot), but importantly also on intra-class level differentiating fine-grained visual differences amongst class instances. 

It is therefore not surprising that traditional few-shot methods, when applied to the {fine-grained} problem, can no longer hold their promises. This is clearly seen in Figure~\ref{fig:1}, where we apply ProtoNet~\cite{NIPS2017_cb8da676} on the ``birds'' dataset~\cite{WelinderEtal2010} -- the model {tends to encourage} lower inter-class variations and higher intra-class variations, which is the very opposite of our goal of fine-grained image classification.

Attempts have been made on adapting traditional few-shot learning methods to the fine-grained scenario. Early attempts~\cite{Huang2021LowRankPA, sun2020few, zhu2020multi} have however focused on devising complex architectural designs which resulted in marginal gains over their vanilla counterparts. It was not until very recently that reconstruction-based methods have gained popularity~\cite{Wertheimer_2021_CVPR, NEURIPS2020_fa28c6cd} and consequently achieved state-of-the art performance. Through specifically engineering for {support-query} feature alignment, they naturally encourage fine-grained transfer. Effects of this can be observed in Figure \ref{fig:1}, where feature regions learned by FRN~\cite{Wertheimer_2021_CVPR} tend to be more fine-grained when compared with ProtoNet~\cite{NIPS2017_cb8da676}.

\begin{figure*}[ht]
  \centering
  \includegraphics[width=0.8\linewidth]{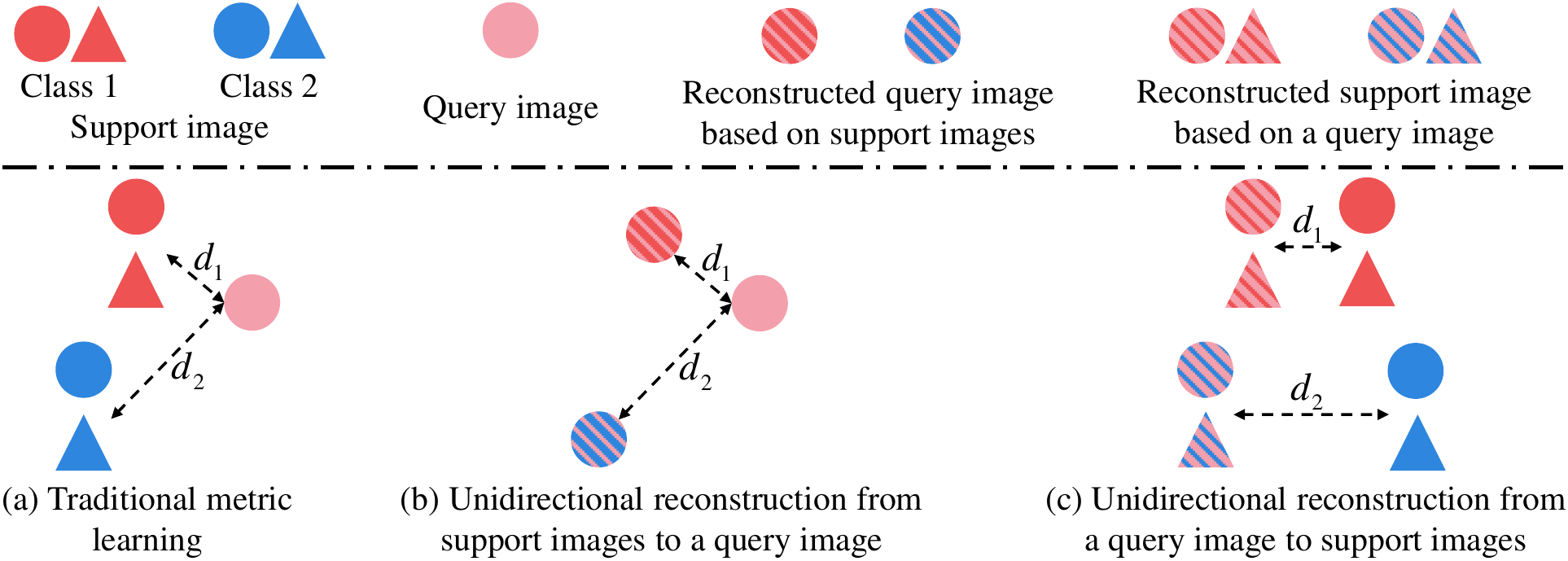}
  \caption{(a) is the traditional metric based method. (b) is the method proposed in~\cite{Wertheimer_2021_CVPR}. (b) + (c) is the method proposed in this paper. (b) can help the model increase the inter-class variations, and (c) can help the model decrease the intra-class variations.}
  \label{fig:1.2}
\end{figure*}

However upon careful inspection, we importantly observe that although the model can focus on more subtle and discriminative regions, the semantic information they represented is nonetheless disparate between any two samples belonging to the same class (e.g., ``Class 1'' FRN focused on different parts of a bird). This seems to suggest that these reconstruction-based methods still suffer from having large intra-class variations, which is counter-productive for overall fine-grained learning (large inter-class variations and small intra-class variations). In other words, such methods only help to increase inter-class variations and do not reduce intra-class variations very well. The reason, we conjecture, is that the existing reconstruction-based methods~\cite{Wertheimer_2021_CVPR, NEURIPS2020_fa28c6cd} (as shown in Figure~\ref{fig:1.2}(b)) works in a unidirectional fashion, i.e., reconstruction only happens one way from support features to query features. 
Consequently, it can only constrain the relationship between reconstructed query features and original query features to increase inter-class variations. This however fails to constrain the relationship between samples within each class in the support set, which is pivotal in decreasing intra-class variations.

As such, in this paper, we for the first time introduce a bi-directional reconstruction mechanism for few-shot fine-grained classification. Instead of using the support set to reconstruct the query set to increase inter-class variations only (as shown in Figure~\ref{fig:1.2}(b)), 
we additionally use the query set to reconstruct the support set to simultaneously reduce intra-class variations (as shown in Figure~\ref{fig:1.2}(c)). This modification might sound overly simple at first sight, it however importantly fulfills both desired learning outcomes for the fine-grained setting -- support to query to encourage large inter-class variations, and query to support to encourage small intra-class variations
(see Section \ref{Ablation} for a detailed explanation).

Our bi-directional feature reconstruction framework mainly contains four modules: (i) a feature extraction module, (ii) a self-reconstruction module, (iii) a mutual bi-directional reconstruction module, and (iv) a distance metric module. (i) and (iv) are common to a reconstruction-based approach~\cite{Wertheimer_2021_CVPR}. In addition to the proposed bi-directional module (iii), we also find use of a self-reconstruction module (ii) to benefit fine-grained feature learning, and works well with the former in terms of increasing inter-class variations and reducing intra-class variations. Ablative studies in Section \ref{Ablation} show both to be effective {and indispensable} for few-shot fine-grained learning. 


In summary, our contributions are three-fold: 
1) We reveal the key problem in current reconstruction-based few-shot fine-grained classification lies with its inability in minimising intra-class variations. 
2) We for the first time propose a bi-directional reconstruction network that simultaneously increase inter-class variations while reducing intra-class variations by way of a mutual {support-query and query-support} reconstruction. 
3) Experimental results and ablative analyses on three fine-grained few-shot image datasets consistently demonstrate the superiority of the proposed method and reveal insights on why the bi-directional approach is effective.

\section{Related works}

\textbf{Metric-Based Few-shot Learning:}
In few-shot learning methods, metric-based methods have received extensive attention owing to their simplicity and efficiency~\cite{Li2021DeepML}. 
{A plethora of earlier works have adopted fixed metric or learnable module to learn a good metric embedding~\cite{NIPS2017_cb8da676, DBLP:conf/nips/VinyalsBLKW16, 8578229}. These methods classify samples according to distance or similarity.} 
Recently, GNN-based few-shot methods~\cite{garcia2018fewshot, Kim_2019_CVPR, Yang_2020_CVPR} adopted graph neural networks (GNN) to model the similarity measurement -- their advantage being that samples have a rich relational structure.
In addition to these classic ones, some metric-based methods for fine-grained image classification have also emerged. 
While DN4~\cite{8953758} proposed a local descriptor-based image-to-class measure, using local features of samples to learn feature metric, {BSNet~\cite{9293172} used a bi-similarity network to learn fewer but more discriminative regions using a combination of two different metrics.} 
NDPNet~\cite{Zhang2021NDPNetAN} designed a feature re-abstraction embedding network that projects the local features into the proposed similarity metric learning network to learn discriminative projection factors. 
Our method adopts a fixed euclidean distance to measure the error between construction features and origin features.

\textbf{Alignment-based Few-shot Learning:}
Alignment-based fine-grained few-shot methods pay more attention to align the spatial positions of objects in images and then classify them based on a similarity measure.
Unlike PARN~\cite{Wu_2019_ICCV} which is a position-aware relational network that aligns similar objects in spatial position and learns more flexible and robust metric capabilities,  
Semantic Alignment Metric Learning (SAML)~\cite{Hao2019CollectAS} aligns the semantically relevant dominant objects in fine-grained images using a collect-and-select strategy.
DeepEMD~\cite{Zhang_2020_CVPR} uses the Earth Mover's Distance to generate the optimal matching flows between two local feature sets and computes the distance between two images based on the optimal matching flows and matching costs. 
In addition to these alignment-based methods above, reconstruction-based methods are also good for aligning spatial positions of objects from different fine-grained images. 
CTX~\cite{NEURIPS2020_fa28c6cd} constructs a novel transformer-based neural network, which can find a coarse spatial relationship between the query and the labelled images, and then compute the distances between spatially-corresponding features to predict the label of samples. 
Alleviating the need for any new modules or large-scale learnable parameters FRN~\cite{Wertheimer_2021_CVPR} obtains the optimal reconstruction weights in the closed form solution to reconstruct query features from support features.
Unlike these existing reconstruction based methods, we introduce bi-directional reconstruction method, that not only reconstructs query samples based on support samples for increasing the inter-class variations, but also reconstructs support samples based on a query sample for reducing the intra-class variations.


\textbf{Attention Mechanism:}
Vaswani et al.~\cite{NIPS2017_3f5ee243} first proposed the self-attention mechanism and then used it to built a new simple network architecture, namely Transformer. 
Besides achieving success in natural language processing~\cite{NIPS2017_3f5ee243, Devlin2019BERTPO, brown2020language}, this architecture also works well in computer vision tasks~\cite{dosovitskiy2021an, hassani2021escaping, Chen2021CrossViTCM}.
Recently, some few-shot learning works have gradually begun to adopt the self-attention mechanism. Han-Jia Ye et al.~\cite{Ye_2020_CVPR} proposed a Few-shot Embedding Adaptation with Transformer (FEAT) for few-shot learning. Here authors constructed set-to-set functions using a variety of approximators and found Transformer to be the most efficient option that {can model interactions between images in a set and hence enable co-adaptation of each image.}
{Unlike FEAT where the transformer structure is only used for support samples, CTX~\cite{NEURIPS2020_fa28c6cd} uses self-attention to calculate the spatial attention weights of query sample and support samples and learn a query-aligned class prototype. It then calculates the distance between the query sample and the aligned class prototype to classify the query sample.}
The proposed few-shot classification method also introduced the self-attention mechanism adopted in the transformer to learn optimal reconstruction weights in our self-reconstruction module and mutual reconstruction module.

\section{Methodology}
In this section, we describe our proposed method, starting with a formulation of the proposed method followed by an overview and an in-depth description of each component.

\subsection{The Problem Formulation}
Given a dataset $D = \{(x_i,y_i), y_i \in Y\}$, we divide it into three parts, that is, $D_{base} = \{(x_i,y_i), y_i \in Y_{base}\}$, $D_{val} = \{(x_i,y_i), y_i \in Y_{val}\}$ and $D_{novel} =\{ (x_i,y_i), y_i \in Y_{novel}\}$, where $x_i$ and  $y_i$ are the original feature vector and class label of the $i^\text{th}$ image, respectively.
The categories of these three parts are disjoint, i.e., $Y_{base} \cap Y_{val} \cap Y_{novel} = \varnothing$, and $Y_{base} \cup Y_{val} \cup Y_{novel} = Y$. 
Few-shot classification aims to improve a $C$-way $K$-shot classification performance on $D_{novel}$ by learning knowledge from $D_{base}$ and $D_{val}$. 
For such a task, there are $C$ classes sampled randomly from $D_{novel}$, and each class only has $K$ {randomly sampled} labelled (support) samples $S$, and $M$ {randomly sampled} unlabelled (query) samples $Q$.

Many few-shot methods adopt the meta-learning paradigm~\cite{Finn2017ModelAgnosticMF, ye2022how}. Specifically, the training process of these methods on $D_{base}$ is the same as the prediction process on $D_{novel}$. Their purpose is to learn the meta-knowledge of learning a category concept based on a few labelled samples.
In the meta-training phase, the few-shot classification model learns the categories of query samples based on few support samples in each task on $D_{base}$. The optimal model is selected by evaluating performance of the {few-shot classification model} on multiple tasks on $D_{val}$. In meta-test phase, the final performance of the optimal model are commonly evaluated by the average accuracy on all sampled tasks of $D_{novel}$.

\subsection{Overview}

\begin{figure*}[!ht]
  \centering
  \includegraphics[width=0.9\linewidth]{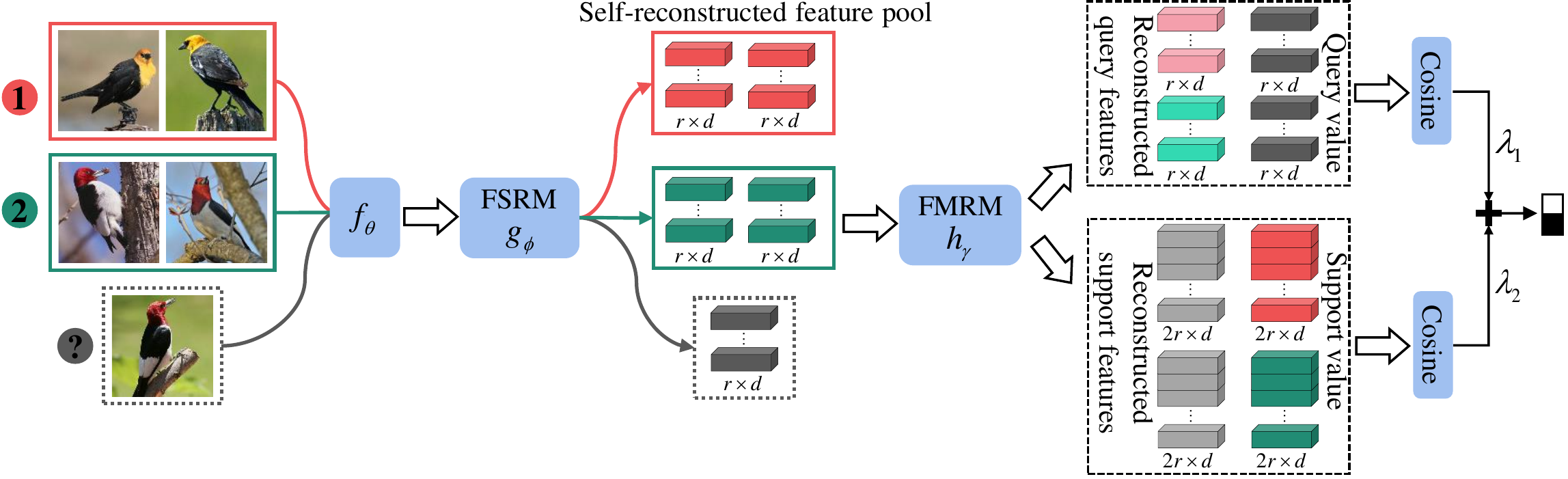}
  \caption{The proposed Bi-Directional feature reconstruction network. FSRM refers to Feature Self-reconstruction Module and FMRM refers to Feature Mutual Reconstruction Module.}
  \label{fig:3.1}
\end{figure*}

Learning subtle and discriminative features is crucial for fine-grained few-shot image classification. Considering that the existing reconstruction-based methods, which mostly reconstruct query features based on support features, fail to reduce intra-class variations adequately, we propose a  bi-directional reconstruction network.

As shown in Figure~\ref{fig:3.1}, our model consists of four modules: 
the first is embedding module $f_{\theta}$ for extracting deep convolutional image features. This can be a traditional convolutional neural network or a residual network.
The second is a feature self-reconstruction module $g_{\phi}$, in which the convolutional features of each image are reconstructed by themselves based on a self-attention mechanism. 
This module can make the similar local features become more similar, while dissimilar ones even more dissimilar, and additionally benefits the following feature mutual reconstruction. 
The third is a feature mutual reconstruction module, $h_{\gamma}$, which reconstructs sample features in a bidirectional form. This module not only uses the support sample to reconstruct the query sample but also reconstructs the support sample from the query sample. Compared with the existing unidirectional reconstruction which only focuses on increasing inter-class variations of features, the bi-directional reconstruction adds another function -- reducing the intra-class variations of features. 
Finally, the fourth is a {Euclidean} metric module, which is in charge of calculating the distance between query sample and reconstructed query sample, as well as support samples and reconstructed support samples. The weighted sum of the two distances are used for classifying query samples.

\subsection{Feature Self-Reconstruction Module (FSRM)}

\begin{figure}[htp]
\centering
  \includegraphics[width=0.85\linewidth]{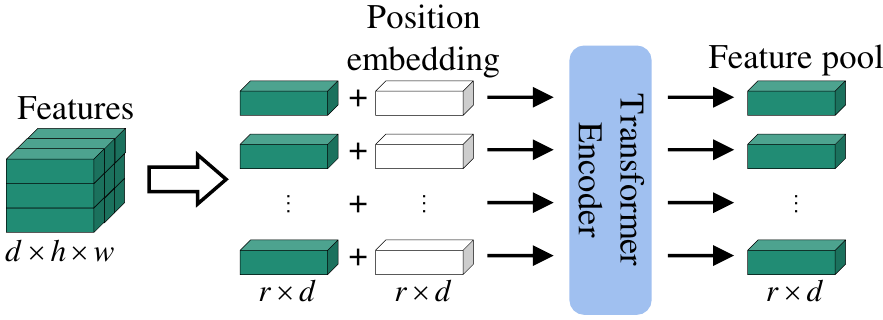}
  \caption{Feature self-reconstruction module.}
  \label{fig:3.2}
\end{figure}

We constructed the feature self-reconstruction module (FSRM), as shown in Figure~\ref{fig:3.2}. For a $C$-way $K$-shot classification task, we input $C \times (K+M)$ samples $x_i$ into the embedding module $f_{\theta}$ to extract features $\hat{x_i} = f_\theta(x_i) \in \mathbb{R}^{d\times h \times w}$, where $d$ is the channel number, $h$ and $w$ represents the height and the width of features, respectively.
Each image feature $\hat{x_i}$ is the input of the FSRM, $g_{\phi}$, and the output of the FSRM is recorded as ${\hat{z_i}} \in \mathbb{R}^{r \times d}$.

First, we reshape the feature $\hat{x_i}$ as $r$ local features in spatial positions $[\hat{x}_i^1, \hat{x}_i^2, \cdots, \hat{x}_i^{r}]$, where $r=h\times w$. The traditional vision transformers~\cite{dosovitskiy2021an} build on standard transformers~\cite{NIPS2017_3f5ee243}, taking the sequence of image patches as input. We {however} compute the summation of the sequence of local features $\hat{x}_i^{j}$ and the corresponding spatial position embedding $E_{pos} \in R^{r \times d}$ as the input of the transformer, i.e.,
  ${z_i} = [\hat{x}_i^1, \hat{x}_i^2, \cdots, \hat{x}_i^{r}] + E_{pos}$,
where $E_{pos}$ adopts sinusoidal position encoding~\cite{NIPS2017_3f5ee243}.

The output of FSRM module is computed based on the standard self-attention operation in Transformer Encoder, and the computing operation is shown as follows:
\begin{equation}
  Attention(Q, K, V) = Softmax(\frac{QK^T}{\sqrt{d_k}})V.
  \label{equ:2}
\end{equation}
Therefore, we can obtain $\hat{z_i}$ as,
\begin{equation}
  \hat{z_i} = Attention(z_i W_{\phi}^Q, z_i W_{\phi}^K, z_i W_{\phi}^V), \hat{z_i} \in \mathbb{R}^{r\times d},
\end{equation}
where $W_{\phi}^Q$, $W_{\phi}^K$, and $W_{\phi}^V$ are a set of learnable weight parameters with $d\times d$ size. Next, the $\hat{z_i}$ are calculated continually by a Layer Normalization ($LN$) and a Multi-Layer Perceptron ($MLP$).
\begin{equation}
  \hat{z_i} = MLP(LN(z_i + \hat{z_i})).
\end{equation}

\subsection{Feature Mutual Reconstruction Module (FMRM)}
\begin{figure}[ht]
  \centering
  \includegraphics[width=0.8\linewidth]{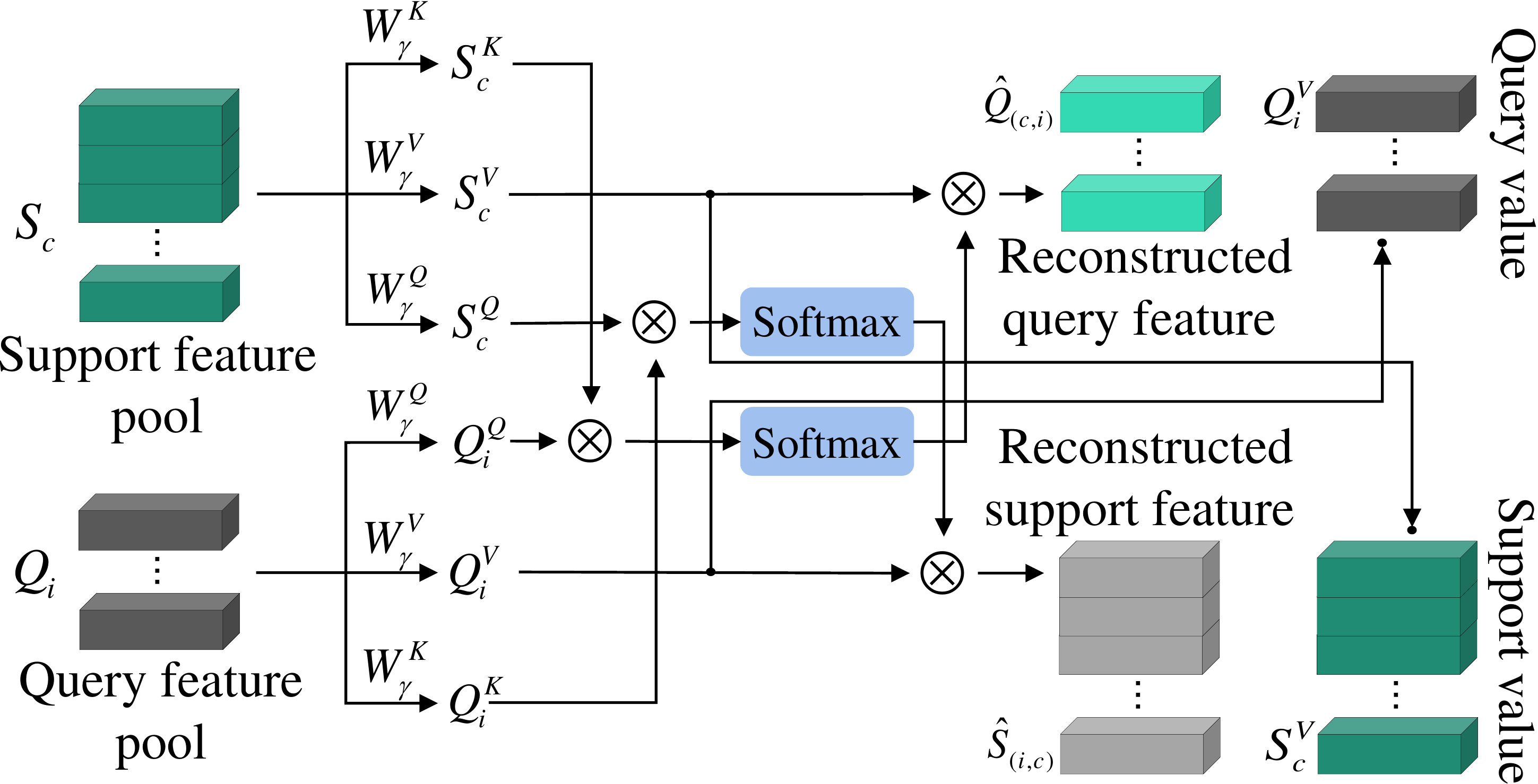}
  \caption{Feature mutual reconstruction module.}
  \label{fig:3.3}
\end{figure}

We propose Feature Mutual Reconstruction Module (FMRM), as shown in Figure~\ref{fig:3.3}, which contains two operations: 
reconstructing support features in one class given a query feature and reconstructing query feature given support features in one class.

For $C$-way $K$-shot classification task, after going through FSRM, we can obtain reconstructed support features of the $c^\text{th}$ class, i.e. $S_c = [\hat{z}_k^c] \in \mathbb{R}^{kr\times d} $, where $k \in [1,\cdots, K]$ {and $c \in [1,\cdots, C]$}, and reconstructed query feature $Q_i =\hat{z_i} \in \mathbb{R}^{r\times d}$, where 
$i \in [1,\cdots, C\times M]$. 
$S_c$ multiplies by weights $W_{\gamma}^Q$, $W_{\gamma}^K$ and $W_{\gamma}^V$,respectively, obtaining $S_c^Q$, $S_c^K$ and $S_c^V$, where $W_{\gamma}^Q, W_{\gamma}^K, W_{\gamma}^V \in \mathbb{R}^{d\times d}$. Similarly, $Q_i$ multiplies by weights $W_{\gamma}^Q$, $W_{\gamma}^K$ and $W_{\gamma}^V$, respectively, obtaining $Q_i^Q$, $Q_i^K$ and $Q_i^V$.

We calculate the reconstructed $i^\text{th}$ query feature $\hat{Q}_{(c,i)}$ from {support features $S_c^V$ in the $c^\text{th}$ class and the reconstructed support features $\hat{S}_{(i,c)}$ in the $c^\text{th}$ class} from $i^\text{th}$ query using the two equations below.

\begin{equation}
\hat{Q}_{(c, i)} = Attention(Q_i^Q, S_c^K, S_c^V), \hat{Q}_{(c, i)} \in \mathbb{R}^{r\times d},
\label{equ:5}
\end{equation}

\begin{equation}
\hat{S}_{(i, c)} = Attention(S_c^Q, Q_i^K, Q_i^V), \hat{S}_{(i, c)}\in \mathbb{R}^{kr\times d}.
\label{equ:6}
\end{equation}
where $Attention(\cdot, \cdot, \cdot)$ is shown in Equation~\ref{equ:2}.

To our best knowledge, the existing reconstruction methods only use unidirectional reconstruction -- using support features to reconstruct the query feature. Building on existing methods, we add the reverse reconstruction shown in Equation~\ref{equ:6} -- using query feature to reconstruct support features.

\subsection{Learning Objectives}

After FMRM, we use the {Euclidean} metric to compute the distance from this query sample $Q_i$ to the support samples in the $c^\text{th}$ class (Figure~\ref{fig:1.2}(b)) as:
\begin{equation}
d_{Q_i \to S_c} = ||Q_i^V - \hat{Q}_{(c, i)}||^2,
\label{equ:7}
\end{equation}
and compute the distance from the support samples in the $c^\text{th}$ class to the query sample $Q_i$ (Figure~\ref{fig:1.2}(c)) as: 
\begin{equation}
d_{S_c \to Q_i} = ||S_c^V - \hat{S}_{(i, c)}||^2.
\label{equ:8}
\end{equation}
The total distance can be obtained by weighted summation of two distances $d_{Q_i \to S_c}$ and $d_{S_c \to Q_i}$, as, 
\begin{equation}
d_i^c = \tau(\lambda_1 d_{Q_i \to S_c} + \lambda_2 d_{S_c \to Q_i}),
\label{equ:9}
\end{equation}
where $\lambda_1$ and $\lambda_2$ are learnable weight parameters, and both of them are initialized as $0.5$. $\tau$ is a learnable temperature factor, following~\cite{Wertheimer_2021_CVPR, Ye_2020_CVPR, Chen2020ANM, Gidaris2018DynamicFV}.  
We normalize $d_i^c$ to obtain $\hat{d_i^c}$ as:
\begin{equation}
\hat{d_i^c} = \frac{e^{-d_i^c}}{\sum_{c=1}^C e^{-d_i^c}}.
\end{equation}
Based on $\hat{d_i^c}$, the total loss in one $C$-way $K$-shot task can be calculated as:
\begin{equation}
\mathcal{L} = -\frac{1}{M \times C} \sum_{i=1}^{M \times C} \sum_{c=1}^C \textbf{1}(y_ i==c) log(\hat{d_i^c}),
\end{equation}
where $\textbf{1}(y_i==c)$ equals $1$ when $y_i$ and $c$ are equal, otherwise $0$.

During the training process, for a $C$-way $K$-shot task on $D_{base}$, we minimize $\mathcal{L}$ to update the proposed network, and repeat this process on all randomly generated tasks.

\section{Experimental results and Analysis}

\subsection{Datasets and Implementation Details}\label{data_imp}
\textbf{Datasets:} 
To evaluate the performance of the proposed method, we selected {three} benchmark fine-grained datasets,
CUB-200-2011 (CUB)~\cite{WelinderEtal2010} is a classic fine-grained image classification dataset. It contains $11,788$ images from $200$ bird species. Following~\cite{Zhang_2020_CVPR,Ye_2020_CVPR}, we crop each image to a human annotated bounding box.
Stanford-Dogs (Dogs)~\cite{KhoslaYaoJayadevaprakashFeiFei_FGVC2011} is a challenging fine-grained image categorization dataset. The dataset includes $20,580$ annotated images of $120$ breeds of dogs from around the world.
Stanford-Cars (Cars)~\cite{6755945} is also a commonly used benchmark dataset for fine-grained image classification. The dataset contains $16,185$ images of $196$ car-types. 
For each dataset, we divided it into $D_{train}$, $D_{val}$ and $D_{test}$. The ratio of $D_{train}$, $D_{val}$ and $D_{test}$ is same as the literature ~\cite{zhu2020multi}, and all images are resized to $84 \times 84$.

\textbf{Implementation Details:}
{We conducted experiments on two widely used backbone architectures: Conv-4 and ResNet-12. The architectures of Conv-4 and ResNet-12 are the same as that of ~\cite{Wertheimer_2021_CVPR, Ye_2020_CVPR}.
We implemented all our experiments on NVIDIA 3090Ti GPUs via Pytorch~\cite{Paszke2019PyTorchAI}.
In our experiments, we train all Conv-4 and ResNet-12 models for $1,200$ epochs using SGD with Nesterov momentum of $0.9$. The initial learning rate is set to $0.1$ and weight decay to 5e-4. Learning rate is decreased by a scaling factor of 10 after every 400 epochs.
For Conv-4 models, we train the proposed models using $30$-way $5$-shot episodes, and test for $1$-shot and $5$-shot episodes. We use $15$ query images per class in both settings. 
Furthermore, for ResNet-12 models we train our proposed model using $15$-way $5$-shot episodes, in order to save memory. 
We employ standard data augmentation, including center crop, random horizontal flip and colour jitter for better training stability. Thereafter, we select the best-performing model based on the validation set, and validate every $20$ epochs. For all experiments, we report the mean accuracy of $10,000$ randomly generated tasks on $D_{test}$ with $95\%$ confidence intervals on the standard $5$-way, $1$-shot and $5$-shot settings.}

\subsection{Comparison with State-of-the-Arts}
\begin{table*}[!ht]
\setlength\tabcolsep{4.8pt}
\centering
\caption{5-way few-shot classification performance on the \textit{CUB}, \textit{Dogs} and \textit{Cars} datasets. {The top block uses Conv-4 backbone and the bottom block uses ResNet-12 backbone.} Methods labeled by $\dag$ denote our implementations.}
\label{tab:1}
\begin{tabular}{ccccccc}
\toprule[1pt]
\multirow{2}{*}{\it{Method}}      
& \multicolumn{2}{c}{\it{CUB}} 
& \multicolumn{2}{c}{\it{Dogs}}  
& \multicolumn{2}{c}{\it{Cars}}  \\ 
                                                   
&\multicolumn{1}{c}{$1$-shot} 
& \multicolumn{1}{c}{$5$-shot}  
& \multicolumn{1}{c}{$1$-shot}  
& \multicolumn{1}{c}{$5$-shot}  
& \multicolumn{1}{c}{$1$-shot} 
& \multicolumn{1}{c}{$5$-shot} \\ \midrule

ProtoNet$\dag$~(NeurIPS 2017)
& 64.82$\pm$0.23       
& 85.74$\pm$0.14          
& 46.66$\pm$0.21    
& 70.77$\pm$0.16  
& 50.88$\pm$0.23
& 74.89$\pm$0.18 \\ 

Relation~(CVPR 2018)
& 63.94$\pm$0.92          
& 77.87$\pm$0.64           
& 47.35$\pm$0.88          
& 66.20$\pm$0.74            
& 46.04$\pm$0.91          
& 68.52$\pm$0.78 \\ 

DN4~(CVPR 2019)
& 57.45$\pm$0.89
& 84.41$\pm$0.58
& 39.08$\pm$0.76
& 69.81$\pm$0.69
& 34.12$\pm$0.68
& {87.47$\pm$0.47} \\

PARN$\dag$~(ICCV 2019)
& {74.43$\pm$0.95}    
& 83.11$\pm$0.67
& 55.86$\pm$0.97
& 68.06$\pm$0.72
& 66.01$\pm$0.94
& 73.74$\pm$0.70 \\

SAML~(ICCV 2019)
& {65.35$\pm$0.65}    
& 78.47$\pm$0.41
& 45.46$\pm$0.36
& 59.65$\pm$0.51
& 61.07$\pm$0.47
& 88.73$\pm$0.49 \\

DeepEMD~(CVPR 2020)
& 64.08$\pm$0.50
& 80.55$\pm$0.71
& 46.73$\pm$0.49
& 65.74$\pm$0.63
& 61.63$\pm$0.27
& 72.95$\pm$0.38 \\

LRPABN~(TMM 2021)
& 63.63$\pm$0.77
& 76.06$\pm$0.58
& 45.72$\pm$0.75
& 60.94$\pm$0.66
& 60.28$\pm$0.76
& 73.29$\pm$0.58 \\

BSNet(D\&C)~(TIP 2021)
& 62.84$\pm$0.95
& 85.39$\pm$0.56
& 43.42$\pm$0.86
& 71.90$\pm$0.68
& 40.89$\pm$0.77
& 86.88$\pm$0.50 \\ 

CTX$\dag$~(NeurrIPS 2020)
& 72.61$\pm$0.21
& 86.23$\pm$0.14
& 57.86$\pm$0.21
& 73.59$\pm$0.16
& 66.35$\pm$0.21
& 82.25$\pm$0.14 \\

FRN$\dag$~(CVPR 2021)
& 74.90$\pm$0.21
& 89.39$\pm$0.12
& 60.41$\pm$0.21
& 79.26$\pm$0.15
& 67.48$\pm$0.22
& 87.97$\pm$0.11 \\ 

FRN+TDM$\dag$~(CVPR 2022)
& 72.01$\pm$0.22        
& 89.05$\pm$0.12          
& 51.57$\pm$0.23
& 75.25$\pm$0.16     
& 65.67$\pm$0.22       
& 86.44$\pm$0.12  \\

FRN+TDM (-noise)$\dag$~(CVPR 2022)
& 76.55$\pm$0.21        
& 90.33$\pm$0.11          
& 62.68$\pm$0.22       
& 79.59$\pm$0.15         
& 71.16$\pm$0.21    
& 89.55$\pm$0.10  \\

Ours 
& \textbf{79.08$\pm$0.20}
& \textbf{92.22$\pm$0.10}
& \textbf{64.74$\pm$0.22}
& \textbf{81.29$\pm$0.14}
& \textbf{75.74$\pm$0.20}
& \textbf{91.58$\pm$0.09} \\ \midrule

ProtoNet$\dag$~~(NeurIPS 2017)
& 81.02$\pm$0.20      
& 91.93$\pm$0.11      
& 73.81$\pm$0.21      
& 87.39$\pm$0.12       
& 85.46$\pm$0.19         
& 95.08$\pm$0.08 \\

CTX$\dag$~(NeurIPS 2020)
& 80.39$\pm$0.20
& 91.01$\pm$0.11
& 73.22$\pm$0.22
& 85.90$\pm$0.13
& 85.03$\pm$0.19
& 92.63$\pm$0.11 \\

DeepEMD$\dag$~(CVPR 2020)
& 75.59$\pm$0.30
& 88.23$\pm$0.18
& 70.38$\pm$0.30
& 85.24$\pm$0.18
& 80.62$\pm$0.26
& 92.63$\pm$0.13  \\

FRN$\dag$~(CVPR 2021)
& 84.30$\pm$0.18   
& 93.34$\pm$0.10         
& {76.76$\pm$0.21}      
& {88.74$\pm$0.12}        
& 88.01$\pm$0.17     
& 95.75$\pm$0.07 \\ 

FRN+TDM$\dag$~(CVPR 2022)
& 85.15$\pm$0.18        
& 93.99$\pm$0.09         
& \textbf{78.02$\pm$0.20}   
& \textbf{89.85$\pm$0.11} 
& 88.92$\pm$0.16
& 96.88$\pm$0.06  \\

FRN+TDM (-noise)$\dag$~(CVPR 2022)
& 84.97$\pm$0.18        
& 93.83$\pm$0.09         
& 77.94$\pm$0.21      
& 89.54$\pm$0.12
& 88.80$\pm$0.16
& 97.02$\pm$0.06  \\ 

Ours                    
& \textbf{85.44$\pm$0.18}   
& \textbf{94.73$\pm$0.09}   
& {76.89$\pm$0.21}   
& 88.27$\pm$0.12 
& \textbf{90.44$\pm$0.15} 
& \textbf{97.49$\pm$0.05} \\ \bottomrule
\end{tabular}
\end{table*}

\begin{table*}[!htp]
\centering
\caption{Ablation studies using only FSRM module or FMRM module.}
\label{tab:3}
\begin{tabular}{ccccccccc}
\toprule[1pt]
  \multirow{2}{*}{\it{Backbone}} 
& 
\multirow{2}{*}{\it{Method}} 
& \multicolumn{2}{c}{\it{CUB}}
& \multicolumn{2}{c}{\it{Dogs}} 
& \multicolumn{2}{c}{\it{Cars}} \\ 

&                             
& \multicolumn{1}{c}{$1$-shot}
& \multicolumn{1}{c}{$5$-shot}
& \multicolumn{1}{c}{$1$-shot}
& \multicolumn{1}{c}{$5$-shot} 
& \multicolumn{1}{c}{$1$-shot}
& \multicolumn{1}{c}{$5$-shot} \\ \midrule
                            
\multirow{4}{*}{Conv-4}
& 
Baseline (ProtoNet)       
& 64.82$\pm$0.23       
& 85.74$\pm$0.14          
& 46.66$\pm$0.21       
& 70.77$\pm$0.16 
& 50.88$\pm$0.23
& 74.89$\pm$0.18 \\ 

& 
(FSRM)             
& 75.37$\pm$0.21
& 88.61$\pm$0.12
& \textbf{65.10$\pm$0.22}
& 79.94$\pm$0.15
& 71.61$\pm$0.22
& 84.70$\pm$0.14 \\ 

& 
(FMRM)              
& 74.92$\pm$0.21
& 89.97$\pm$0.11
& 61.28$\pm$0.21
& 80.07$\pm$0.14 
& 70.22$\pm$0.21       
& 88.45$\pm$0.11 \\ 

&
(FSRM+FMRM)
& \textbf{79.08$\pm$0.20}
& \textbf{92.22$\pm$0.10}
& 64.74$\pm$0.22
& \textbf{81.29$\pm$0.14} 
& \textbf{75.74$\pm$0.20}
& \textbf{91.58$\pm$0.09} \\ \midrule

\multirow{4}{*}{ResNet-12}
& 
Baseline (ProtoNet)         
& 81.02$\pm$0.20
& 91.93$\pm$0.11
& 73.81$\pm$0.21
& 87.39$\pm$0.12 
& 85.46$\pm$0.19         
& 95.08$\pm$0.08 \\ 

& 
(FSRM)
& 82.53$\pm$0.19
& 92.43$\pm$0.10
& 75.64$\pm$0.21
& 87.44$\pm$0.12 
& 85.95$\pm$0.18       
& 94.44$\pm$0.08 \\
                            
& 
(FMRM)
& 84.33$\pm$0.18
& 94.25$\pm$0.09
& 76.29$\pm$0.21
& \textbf{89.06$\pm$0.11} 
& 89.62$\pm$0.15       
& 97.45$\pm$0.05 \\

& 
(FSRM+FMRM)
& \textbf{85.44$\pm$0.18}
& \textbf{94.73$\pm$0.09}   
& \textbf{76.89$\pm$0.21}
& 88.27$\pm$0.12 
& \textbf{90.44$\pm$0.15} 
& \textbf{97.49$\pm$0.05}  \\ \bottomrule
\end{tabular}
\end{table*}

To validate the efficiency of our method for fine-grained few-shot image classification, we conducted experiments on the {three} fine-grained image classification datasets discussed earlier.
The results of Relation~\cite{8578229}, DN4~\cite{8953758} and BSNet~\cite{9293172} are from literature BSNet~\cite{9293172}, and the results of SAML~\cite{Hao2019CollectAS} and DeepEMD~\cite{Zhang_2020_CVPR} are from DLG~\cite{Cao2022AFF}, and the results of LRPABN~\cite{Huang2021LowRankPA} are from MattML~\cite{zhu2020multi}. The results of methods marked with $\dag$, such as ProtoNet~\cite{NIPS2017_cb8da676}, PARN~\cite{Wu_2019_ICCV}, CTX~\cite{NEURIPS2020_fa28c6cd}, FRN~\cite{Wertheimer_2021_CVPR}, FRN+TDM~\cite{lee2022task} and DeepEMD~\cite{Zhang_2020_CVPR} are obtained via the code officially provided by the author, which is replaced by the dataset used in this paper.

\begin{table*}[!htp]
\centering
\caption{Ablation on reconstruction designs of FMRM.}
\label{tab:4}
\begin{tabular}{cccccccc}
\toprule[1pt]

\multirow{2}{*}{\it{Backbone}} 
& \multirow{2}{*}{\it{Method}} 
& \multicolumn{2}{c}{\it{CUB}}                                  
& \multicolumn{2}{c}{\it{Dogs}} 
& \multicolumn{2}{c}{\it{Cars}} \\  

&                             
& \multicolumn{1}{c}{$1$-shot}    
& \multicolumn{1}{c}{$5$-shot}    
& \multicolumn{1}{c}{$1$-shot}     
& \multicolumn{1}{c}{$5$-shot} 
& \multicolumn{1}{c}{$1$-shot}     
& \multicolumn{1}{c}{$5$-shot} \\ \midrule

\multirow{4}{*}{Conv-4}
& 
Baseline (ProtoNet)     
& 64.82$\pm$0.23
& 85.74$\pm$0.14
& 46.66$\pm$0.21
& 70.77$\pm$0.16 
& 50.88$\pm$0.23
& 74.89$\pm$0.18 \\ 

& 
Ours (Q$\to$S)
& \textbf{79.88$\pm$0.20}
& 91.76$\pm$0.11
& \textbf{65.26$\pm$0.22}
& 80.81$\pm$0.14 
& 75.61$\pm$0.20       
& 90.49$\pm$0.10 \\ 

& 
Ours (S$\to$Q)         
& 76.54$\pm$0.21
& 88.03$\pm$0.14
& 64.39$\pm$0.22
& 78.36$\pm$0.15 
& 72.71$\pm$0.22       
& 85.11$\pm$0.14 \\ 
                            
& 
Ours (Mutual)
&  79.08$\pm$0.20
& \textbf{92.22$\pm$0.10}
&  64.74$\pm$0.22
& \textbf{81.29$\pm$0.14} 
& \textbf{75.74$\pm$0.20}
& \textbf{91.58$\pm$0.09} \\ \midrule
                            
\multirow{4}{*}{ResNet-12}
& 
Baseline (ProtoNet)               
& 81.02$\pm$0.20
& 91.93$\pm$0.11
& 73.81$\pm$0.21
& 87.39$\pm$0.12 
& 85.46$\pm$0.19         
& 95.08$\pm$0.08 \\ 

& 
Ours (Q$\to$S)          
& 83.72$\pm$0.19
& 93.31$\pm$0.09
& 76.50$\pm$0.21
& 87.95$\pm$0.12 
& 87.37$\pm$0.17       
& 95.10$\pm$0.08 \\
                            
& 
Ours (S$\to$Q) 
& 81.72$\pm$0.19
& 90.83$\pm$0.11
& 75.62$\pm$0.22
& 86.47$\pm$0.13 
& 85.90$\pm$0.18       
& 93.17$\pm$0.10 \\ 
                            
& 
Ours (Mutual)
& \textbf{85.44$\pm$0.18}
& \textbf{94.73$\pm$0.09}
& \textbf{76.89$\pm$0.21}
& \textbf{88.27$\pm$0.12} 
& \textbf{90.44$\pm$0.15} 
& \textbf{97.49$\pm$0.05} \\ \bottomrule
\end{tabular}
\end{table*}

\begin{figure*}[ht]
  \centering
  \includegraphics[width=0.8\linewidth]{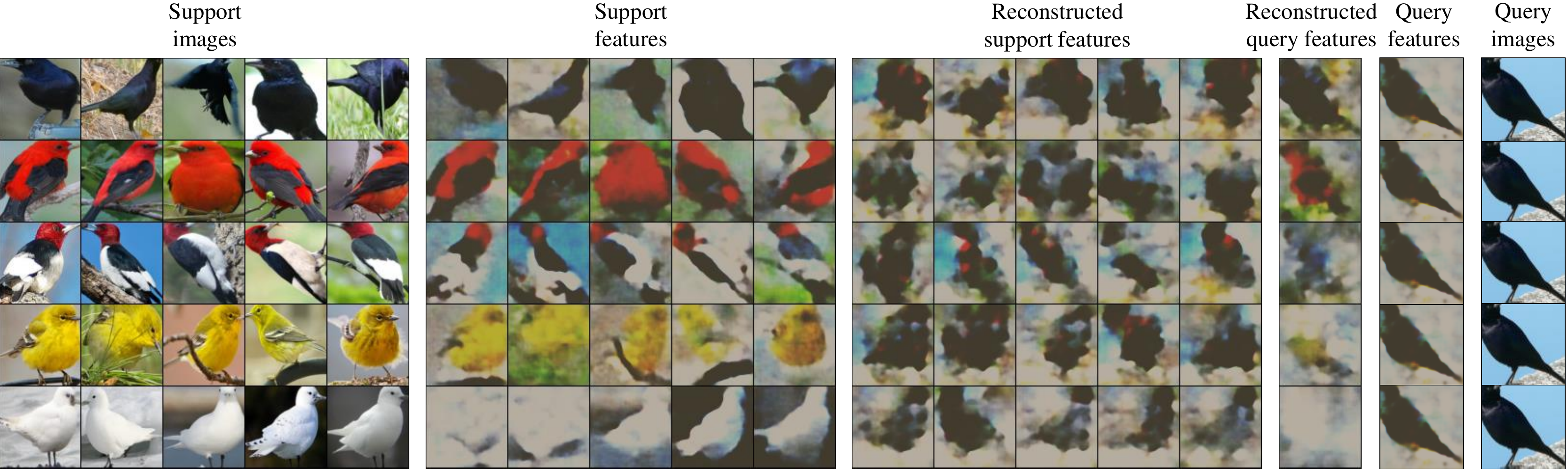}
  \caption{Recovered images of different features by our method for the CUB dataset. } 
  \label{fig:visualize}
\end{figure*}

We use Conv-4 and ResNet-12 as the backbone of all compared methods and test $5$-way $1$-shot and $5$-way $5$-shot classification performance. 
As seen from Table~\ref{tab:1}, the proposed method achieves highest accuracy on all three datasets when the Conv-4 is adopted.
Apart from the result on the {Dogs}~\cite{KhoslaYaoJayadevaprakashFeiFei_FGVC2011} dataset where performance falls slightly behind the FRN+TDM method when the ResNet-12 is adopted, our method achieves highest accuracy.

In a nutshell, compared with other newly proposed methods, our method achieves stable and excellent performance on the three fine-grained image datasets on both $5$-way $1$-shot and $5$-way $5$-shot classification tasks, majorly owing to our network structure. The two modules FSRM and FMRM, conditions the model to learn the subtle and discriminative features for fine-grained classification.

\subsection{Ablation Study}\label{Ablation}

To further justify the design choices of our method and model components towards efficiency and accuracy, we perform a few ablation studies on three datasets using both Conv-4 and ResNet-12 as backbone networks.

\textbf{The Effectiveness of FSRM and FMRM:} We compare our method's efficacy by removing components in a strip-down fashion. In other words, we conduct experiments and report in Table~\ref{tab:3}, where we remove FSRM module (\textit{FMRM}), FMRM module (\textit{FSRM}), and then both which becomes equivalent to ProtoNet~\cite{NIPS2017_cb8da676} (\textit{Baseline (ProtoNet)}).
{Evidently, performance improves further in our method where both FSRM and FMRM are used together (\textit{FSRM+FMRM}). Therefore, FSRM and FMRM modules are indispensable and complementary.}

\textbf{The Effectiveness of Mutual Reconstruction in FMRM:} 
For our method, we remove reconstruction of support samples based on query samples in FMRM, i.g., $\lambda_2$ becoming $0$ in Equation~\ref{equ:9}, which is marked as Ours ($Q \to S$). And we remove reconstruction of query samples based on support samples, i.g., $\lambda_1$ becoming $0$ in Equation~\ref{equ:9}, which is marked as Ours ($S\to Q$). 
{As per Table~\ref{tab:4}, both unidirectional reconstruction methods score lower than the proposed feature mutual reconstruction method (FMRM) in most cases. Therefore, it can be concluded that the design of the Feature Mutual Reconstruction Module (FMRM) is reasonable and effective.}

\subsection{Visualization Analysis}

To demonstrate the efficiency of our proposed network better, we recovered the original and reconstructed features. We trained an inverse ResNet as a decoder, the input of which is the feature value in the mutual reconstruction module, and the output of which is a $3 \times 84 \times 84$ recovered image. In the training process, we use an Adam optimizer with an initial learning rate of $0.01$, set batch size as $200$ and train for $1,000$ epochs, with an L1 loss to measure the prediction error.

As per Figure ~\ref{fig:visualize}, the left-most block shows the support images of $5$ classes, each of which occupies one row and contains $5$ images. The right-most column is a query image, which we copied $5$ times simply for convenient comparison and aesthetics.
The second block from left is the recovered images of support features $S_c^V$, whereas second column from the right is the recovered images of the $i^\text{th}$ query features $Q_i^V$. 
The third block from left is the recovered images of reconstructed support feature $\hat{S}_{(i,c)}$ based on $i^\text{th}$ query feature, while that from the right is the recovered images of reconstructed query features $\hat{Q}_{(c,i)}$ based on each $c$ class support feature.

From the first and second {blocks} (first being left-most), it can be seen that the decoder we trained is capable of recovering images from features. For the third {and fourth ones}, it can be seen that if  a query sample is used to reconstruct the support samples which has same class as the query sample, the reconstructed features will be similar to the original support features. However if we use a query sample to reconstruct the support samples which has a class different from the query sample, the reconstructed features will be quite different from the original support features. Likewise, when we use support samples to reconstruct the query samples, similar results are obtained. This indicates that the proposed network can effectively alleviate inaccurate similarity measure of the unaligned fine-grained images.

\section{Conclusion}
\label{Conclusion}
In this paper, we proposed a bi-directional feature reconstruction network for few-shot fine-grained image classification. Our major contribution is a mutual reconstruction module that works in both directions, i.e., support to test and test to support. Compared to the existing reconstruction-based methods, the proposed method can achieve larger inter-class variations and lower the intra-class variations which is crucial to fine-grained learning. 
Extensive experiments show that the proposed network can perform well on {three} fine-grained image datasets consistently, competing strongly and at times surpassing contemporary state-of-the-arts.


\bibliography{reference}

\begin{thebibliography}{37}
\providecommand{\natexlab}[1]{#1}

\bibitem[{Brown et~al.(2020)Brown, Mann, Ryder, Subbiah, Kaplan, Dhariwal,
  Neelakantan, Shyam, Sastry, Askell et~al.}]{brown2020language}
Brown, T.~B.; Mann, B.; Ryder, N.; Subbiah, M.; Kaplan, J.; Dhariwal, P.;
  Neelakantan, A.; Shyam, P.; Sastry, G.; Askell, A.; et~al. 2020.
\newblock Language Models are Few-Shot Learners.
\newblock \emph{ArXiv}.

\bibitem[{Cao et~al.(2022)Cao, Wang, Zhang, Zheng, and Li}]{Cao2022AFF}
Cao, S.; Wang, W.; Zhang, J.; Zheng, M.; and Li, Q. 2022.
\newblock A Few-shot Fine-grained Image Classification Method Leveraging Global
  and Local Structures.
\newblock \emph{International Journal of Machine Learning and Cybernetics}.

\bibitem[{Chen, Fan, and Panda(2021)}]{Chen2021CrossViTCM}
Chen, C.-F.; Fan, Q.; and Panda, R. 2021.
\newblock CrossViT: Cross-Attention Multi-Scale Vision Transformer for Image
  Classification.
\newblock In \emph{ICCV}.

\bibitem[{Chen et~al.(2020)Chen, Wang, Liu, Xu, and Darrell}]{Chen2020ANM}
Chen, Y.; Wang, X.; Liu, Z.; Xu, H.; and Darrell, T. 2020.
\newblock A New Meta-Baseline for Few-Shot Learning.
\newblock \emph{ArXiv}.

\bibitem[{Doersch, Gupta, and Zisserman(2020)}]{NEURIPS2020_fa28c6cd}
Doersch, C.; Gupta, A.; and Zisserman, A. 2020.
\newblock CrossTransformers: Spatially-aware Few-shot Transfer.
\newblock In \emph{NerulPS}.

\bibitem[{Dosovitskiy et~al.(2021)Dosovitskiy, Beyer, Kolesnikov, Weissenborn,
  Zhai, Unterthiner, Dehghani, Minderer, Heigold, Gelly
  et~al.}]{dosovitskiy2021an}
Dosovitskiy, A.; Beyer, L.; Kolesnikov, A.; Weissenborn, D.; Zhai, X.;
  Unterthiner, T.; Dehghani, M.; Minderer, M.; Heigold, G.; Gelly, S.; et~al.
  2021.
\newblock An Image is Worth 16x16 Words: Transformers for Image Recognition at
  Scale.
\newblock In \emph{ICLR}.

\bibitem[{Finn, Abbeel, and Levine(2017)}]{Finn2017ModelAgnosticMF}
Finn, C.; Abbeel, P.; and Levine, S. 2017.
\newblock Model-agnostic Meta-learning for Fast Adaptation of Deep Networks.
\newblock In \emph{ICML}.

\bibitem[{Gidaris and Komodakis(2018)}]{Gidaris2018DynamicFV}
Gidaris, S.; and Komodakis, N. 2018.
\newblock Dynamic Few-Shot Visual Learning Without Forgetting.
\newblock In \emph{CVPR}.

\bibitem[{Hao et~al.(2019)Hao, He, Cheng, Wang, Cao, and
  Tao}]{Hao2019CollectAS}
Hao, F.; He, F.; Cheng, J.; Wang, L.; Cao, J.; and Tao, D. 2019.
\newblock Collect and Select: Semantic Alignment Metric Learning for Few-shot
  Learning.
\newblock In \emph{ICCV}.

\bibitem[{Hassani et~al.(2021)Hassani, Walton, Shah, Abuduweili, Li, and
  Shi}]{hassani2021escaping}
Hassani, A.; Walton, S.; Shah, N.; Abuduweili, A.; Li, J.; and Shi, H. 2021.
\newblock Escaping the Big Data Paradigm with Compact Transformers.
\newblock \emph{ArXiv}.

\bibitem[{Huang et~al.(2021)Huang, Zhang, Zhang, Xu, and
  Wu}]{Huang2021LowRankPA}
Huang, H.; Zhang, J.; Zhang, J.; Xu, J.; and Wu, Q. 2021.
\newblock Low-rank Pairwise Alignment Bilinear Network for Few-shot
  Fine-grained Image Classification.
\newblock \emph{IEEE Transactions on Multimedia}.

\bibitem[{Kenton and Toutanova(2019)}]{Devlin2019BERTPO}
Kenton, J. D. M.-W.~C.; and Toutanova, L.~K. 2019.
\newblock BERT: Pre-training of Deep Bidirectional Transformers for Language
  Understanding.
\newblock In \emph{NAACL-HLT}.

\bibitem[{Khosla et~al.(2011)Khosla, Jayadevaprakash, Yao, and
  Li}]{KhoslaYaoJayadevaprakashFeiFei_FGVC2011}
Khosla, A.; Jayadevaprakash, N.; Yao, B.; and Li, F.-F. 2011.
\newblock Novel Dataset for Fine-grained Image Categorization: Stanford Dogs.
\newblock In \emph{CVPR workshops}.

\bibitem[{Kim et~al.(2019)Kim, Kim, Kim, and Yoo}]{Kim_2019_CVPR}
Kim, J.; Kim, T.; Kim, S.; and Yoo, C.~D. 2019.
\newblock Edge-Labeling Graph Neural Network for Few-shot Learning.
\newblock In \emph{CVPR}.

\bibitem[{Krause et~al.(2013)Krause, Stark, Deng, and Fei-Fei}]{6755945}
Krause, J.; Stark, M.; Deng, J.; and Fei-Fei, L. 2013.
\newblock 3D Object Representations for Fine-Grained Categorization.
\newblock In \emph{ICCV Workshops}.

\bibitem[{Lee, Moon, and Heo(2022)}]{lee2022task}
Lee, S.; Moon, W.; and Heo, J.-P. 2022.
\newblock Task Discrepancy Maximization for Fine-Grained Few-Shot
  Classification.
\newblock In \emph{CVPR}.

\bibitem[{Li et~al.(2019)Li, Wang, Xu, Huo, Gao, and Luo}]{8953758}
Li, W.; Wang, L.; Xu, J.; Huo, J.; Gao, Y.; and Luo, J. 2019.
\newblock Revisiting Local Descriptor Based Image-To-Class Measure for Few-Shot
  Learning.
\newblock In \emph{CVPR}.

\bibitem[{Li et~al.(2021{\natexlab{a}})Li, Wu, Sun, Ma, Cao, and Xue}]{9293172}
Li, X.; Wu, J.; Sun, Z.; Ma, Z.; Cao, J.; and Xue, J.-H. 2021{\natexlab{a}}.
\newblock BSNet: Bi-Similarity Network for Few-shot Fine-grained Image
  Classification.
\newblock \emph{IEEE Transactions on Image Processing}.

\bibitem[{Li et~al.(2021{\natexlab{b}})Li, Yang, Ma, and Xue}]{Li2021DeepML}
Li, X.; Yang, X.; Ma, Z.; and Xue, J.-H. 2021{\natexlab{b}}.
\newblock Deep Metric Learning for Few-shot Image Classification: A Selective
  Review.
\newblock \emph{ArXiv}.

\bibitem[{Muhammad and Yeasin(2020)}]{muhammad2020eigen}
Muhammad, M.~B.; and Yeasin, M. 2020.
\newblock Eigen-cam: Class Activation Map Using Principal Components.
\newblock In \emph{IJCNN}.

\bibitem[{Paszke et~al.(2019)Paszke, Gross, Massa, Lerer, Bradbury, Chanan,
  Killeen, Lin, Gimelshein, Antiga et~al.}]{Paszke2019PyTorchAI}
Paszke, A.; Gross, S.; Massa, F.; Lerer, A.; Bradbury, J.; Chanan, G.; Killeen,
  T.; Lin, Z.; Gimelshein, N.; Antiga, L.; et~al. 2019.
\newblock PyTorch: An Imperative Style, High-Performance Deep Learning Library.
\newblock In \emph{NeurIPS}.

\bibitem[{Satorras and Estrach(2018)}]{garcia2018fewshot}
Satorras, V.~G.; and Estrach, J.~B. 2018.
\newblock Few-shot Learning with Graph Neural Networks.
\newblock In \emph{ICLR}.

\bibitem[{Snell, Swersky, and Zemel(2017)}]{NIPS2017_cb8da676}
Snell, J.; Swersky, K.; and Zemel, R. 2017.
\newblock Prototypical Networks for Few-shot Learning.
\newblock In \emph{NerulPS}.

\bibitem[{Sun et~al.(2020)Sun, Xv, Dong, Zhou, Chen, and Li}]{sun2020few}
Sun, X.; Xv, H.; Dong, J.; Zhou, H.; Chen, C.; and Li, Q. 2020.
\newblock Few-shot Learning for Domain-specific Fine-grained Image
  Classification.
\newblock \emph{IEEE Transactions on Industrial Electronics}.

\bibitem[{Sung et~al.(2018)Sung, Yang, Zhang, Xiang, Torr, and
  Hospedales}]{8578229}
Sung, F.; Yang, Y.; Zhang, L.; Xiang, T.; Torr, P.~H.; and Hospedales, T.~M.
  2018.
\newblock Learning to Compare: Relation Network for Few-Shot Learning.
\newblock In \emph{CVPR}.

\bibitem[{Vaswani et~al.(2017)Vaswani, Shazeer, Parmar, Uszkoreit, Jones,
  Gomez, Kaiser, and Polosukhin}]{NIPS2017_3f5ee243}
Vaswani, A.; Shazeer, N.; Parmar, N.; Uszkoreit, J.; Jones, L.; Gomez, A.~N.;
  Kaiser, L.~u.; and Polosukhin, I. 2017.
\newblock Attention is All you Need.
\newblock In \emph{NerulPS}.

\bibitem[{Vinyals et~al.(2016)Vinyals, Blundell, Lillicrap, Kavukcuoglu, and
  Wierstra}]{DBLP:conf/nips/VinyalsBLKW16}
Vinyals, O.; Blundell, C.; Lillicrap, T.; Kavukcuoglu, K.; and Wierstra, D.
  2016.
\newblock Matching Networks for One Shot Learning.
\newblock In \emph{NerulPS}.

\bibitem[{Wah et~al.(2011)Wah, Branson, Welinder, Perona, and
  Belongie}]{WelinderEtal2010}
Wah, C.; Branson, S.; Welinder, P.; Perona, P.; and Belongie, S.~J. 2011.
\newblock The Caltech-UCSD Birds-200-2011 Dataset.
\newblock Technical report, California Institute of Technology.

\bibitem[{Wei et~al.(2021)Wei, Song, {Mac Aodha}, Wu, Peng, Tang, Yang, and
  Belongie}]{3401454f29a841a29142c97833857e78}
Wei, X.-S.; Song, Y.-Z.; {Mac Aodha}, O.; Wu, J.; Peng, Y.; Tang, J.; Yang, J.;
  and Belongie, S. 2021.
\newblock Fine-Grained Image Analysis with Deep Learning: A Survey.
\newblock \emph{IEEE Transactions on Pattern Analysis and Machine
  Intelligence}.

\bibitem[{Wertheimer, Tang, and Hariharan(2021)}]{Wertheimer_2021_CVPR}
Wertheimer, D.; Tang, L.; and Hariharan, B. 2021.
\newblock Few-shot Classification With Feature Map Reconstruction Networks.
\newblock In \emph{CVPR}.

\bibitem[{Wu et~al.(2019)Wu, Li, Guo, and Jia}]{Wu_2019_ICCV}
Wu, Z.; Li, Y.; Guo, L.; and Jia, K. 2019.
\newblock PARN: Position-aware Relation Networks for Few-shot Learning.
\newblock In \emph{ICCV}.

\bibitem[{Yang et~al.(2020)Yang, Li, Zhang, Zhou, Zhou, and
  Liu}]{Yang_2020_CVPR}
Yang, L.; Li, L.; Zhang, Z.; Zhou, X.; Zhou, E.; and Liu, Y. 2020.
\newblock DPGN: Distribution Propagation Graph Network for Few-shot Learning.
\newblock In \emph{CVPR}.

\bibitem[{Ye and Chao(2022)}]{ye2022how}
Ye, H.-J.; and Chao, W.-L. 2022.
\newblock How to Train Your MAML to Excel in Few-shot Classification.
\newblock In \emph{ICLR}.

\bibitem[{Ye et~al.(2020)Ye, Hu, Zhan, and Sha}]{Ye_2020_CVPR}
Ye, H.-J.; Hu, H.; Zhan, D.-C.; and Sha, F. 2020.
\newblock Few-shot Learning via Embedding Adaptation With Set-to-Set Functions.
\newblock In \emph{CVPR}.

\bibitem[{Zhang et~al.(2020)Zhang, Cai, Lin, and Shen}]{Zhang_2020_CVPR}
Zhang, C.; Cai, Y.; Lin, G.; and Shen, C. 2020.
\newblock DeepEMD: Few-shot Image Classification With Differentiable Earth
  Mover's Distance and Structured Classifiers.
\newblock In \emph{CVPR}.

\bibitem[{Zhang et~al.(2021)Zhang, Liu, Xue, Gao, and Sun}]{Zhang2021NDPNetAN}
Zhang, W.; Liu, X.; Xue, Z.; Gao, Y.; and Sun, C. 2021.
\newblock NDPNet: A Novel Non-linear Data Projection Network for Few-shot
  Fine-grained Image Classification.
\newblock \emph{ArXiv}.

\bibitem[{Zhu, Liu, and Jiang(2020)}]{zhu2020multi}
Zhu, Y.; Liu, C.; and Jiang, S. 2020.
\newblock Multi-attention Meta Learning for Few-shot Fine-grained Image
  Recognition.
\newblock In \emph{IJCAI}.

\end{thebibliography}

\end{document}